\documentclass[letterpaper, 10 pt, conference]{ieeeconf}

\IEEEoverridecommandlockouts                              

\newcommand{\secref}[1]{Section~\ref{#1}}
\newcommand{\figref}[1]{Fig.~\ref{#1}}

\usepackage{graphicx} 
\usepackage{amsmath, amssymb}
\usepackage{mathtools}
\usepackage{color, soul}
\usepackage{commath}
\usepackage{dblfloatfix}
\DeclareMathOperator{\EX}{\mathbb{E}}
\usepackage{balance}

\title{\LARGE \bf
Inclined Quadrotor Landing using Deep Reinforcement Learning
}

\author{Jacob E. Kooi$^{1}$ and Robert Babu\v{s}ka$^{2}$
\thanks{$^{1}$Jacob E. Kooi is with the Department of Cognitive Robotics and Delft Center for Systems and Control,
        Delft University of Technology, 2628 CD Delft, The Netherlands
        {\tt\small jacobkooi92@gmail.com}}%
\thanks{$^{2}$Robert Babu\v{s}ka with the Department of Cognitive Robotics, Delft University of Technology, 2628 CD Delft, The Netherlands and with the Czech Institute of Informatics, Robotics, and Cybernetics, Czech Technical University in
Prague, Czech Republic
{\tt\small r.babuska@tudelft.nl}}%
}

\begin{document}

\maketitle
\thispagestyle{empty}
\pagestyle{empty}

\begin{abstract}
Landing a quadrotor on an inclined surface is a challenging maneuver. The final state of any inclined landing trajectory is not an equilibrium, which precludes the use of most conventional control methods. We propose a deep reinforcement learning approach to design an autonomous landing controller for inclined surfaces. Using the proximal policy optimization (PPO) algorithm with sparse rewards and a tailored curriculum learning approach, an inclined landing policy can be trained in simulation in less than 90 minutes on a standard laptop. The policy then directly runs on a real Crazyflie 2.1 quadrotor and successfully performs real inclined landings in a flying arena. A single policy evaluation takes approximately 2.5\,ms, which makes it suitable for a future embedded implementation on the quadrotor.

\end{abstract}

\section{Introduction}

Modern quadrotors are agile and can perform complex tasks in difficult-to-reach places. Quadrotor flight and maneuvers are commonly controlled by proportional integral derivative (PID) control or model predictive control (MPC). Although these methods are adequate for set-point or trajectory tracking, they fall short when it comes to more complicated maneuvers that exceed the linearization range or require long prediction horizons. One such maneuver is the landing on an inclined surface, which is relevant for applications like delivery, maintenance, or surveillance. To facilitate a safe inclined landing, the final attitude of the quadrotor must match the slope of the
landing platform.
The  final state of the landing trajectory is not an equilibrium, which presents a challenge for the control design. Owing to the under-actuated nature of the system, the landing trajectory can be long and complex, with an initial motion away from the landing location. This complicates the use of standard control methods like MPC with a fixed prediction horizon and quadratic cost function.

Recent advances in deep reinforcement learning (DRL) with continuous action spaces have made this approach suitable also for quadrotor control \cite{QuadRL2017Hwangbo, UAVRL, modelbasedquadlambert, Honigdeeprlcrazyflie}, including landing controllers \cite{Neural_Lander, Landing_Guidance_UAV_vision, DRL_Landing_Marker, polvara_decklanding, UAV_Marker_Lander_Reality, Rodriguez_platform_DDPG_Landing,RL_landing_higherlevelcommands_LSTP}. However, no results have yet been reported for inclined landing.
In this paper, we develop a DRL approach to the inclined landing problem and validate it in simulations and  real lab experiments with the Crazyflie 2.1 Nano-UAV. To the best of our knowledge, this is the first deep-learning-based controller for inclined landing applied to a real quadcopter. More specifically, our contributions are:
\begin{itemize}
\item We develop two fast Gym-based \cite{Gym} simulation environments for the Crazyflie 2.1 Nano-UAV.\footnote[3]{https://github.com/Jacobkooi/InclinedDroneLander.git} One three-dimensional environment can be used with any compatible DRL algorithm to train set-point tracking. The other two-dimensional environment, restricted to the vertical $xz$-plane, can be used with an on-policy algorithm to train the inclined landing. The resulting policies adequately transfer to the real Crazyflie.

\item Building upon the state-of-the-art model-free proximal policy optimization (PPO) algorithm \cite{PPOSchulmanWDRK17}, we propose a powerful curriculum learning \cite{curriculum} approach to facilitate convergence when using sparse rewards, without the need for applying a multi-goal setting like in hindsight experience replay \cite{HER17} or iterated supervised learning \cite{Iterated_supervised_learning}.

\item We test the trained policy network in simulation and then deploy it to the real Crazyflie quadrotor to demonstrate the actual inclined landing in an indoor flying arena.\footnote[4]{https://youtu.be/pJ6vVs0BsB8}
\end{itemize}
The remainder of the paper is structured as follows. We first give an overview of the related work in \secref{sec:related}. The dynamic quadrotor model used for simulation and training is described in \secref{model}. Next, \secref{method} presents the DRL simulation framework that is used to train inclined landing and set-point tracking. \secref{experiments} describes the simulation and lab setup and presents the experimental results. \secref{discussion} contains a discussion of the results and in \secref{reality}, the conclusions and limitations of this work are given, along with proposals for future work.

\section{Related Work}\label{sec:related}

Deep reinforcement learning methods have been applied to a variety of quadrotor control problems, including hovering \cite{modelbasedquadlambert}, attitude control \cite{UAVRL}, set-point tracking, and disturbance recovery \cite{QuadRL2017Hwangbo,Honigdeeprlcrazyflie}.
Specifically for landing, a deep neural network was employed to learn higher-order interactions to stabilize the near-ground behavior of a nonlinear quadrotor controller \cite{Neural_Lander}. A deep Q-learning network (DQN) was used to detect a marker symbol and perform a landing by using a downward-facing low-resolution camera \cite{DRL_Landing_Marker, polvara_decklanding, UAV_Marker_Lander_Reality}. The work in \cite{polvara_decklanding} considers platform inclination, but only in the context of more involved  visual recognition, while the landing is still horizontal. Least-squares policy iteration (LSPI) was employed to autonomously land on a marker \cite{RL_landing_higherlevelcommands_LSTP} and the deep deterministic policy gradient (DDPG) algorithm \cite{DDPGLillicrap} was used to navigate a descending quadrotor to land on a moving platform \cite{Rodriguez_platform_DDPG_Landing}. Finally, the work in \cite{Landing_Guidance_UAV_vision} involved a convolutional neural network to estimate the heading angle to aid UAV landing in the case of sensor failure. However, none of the approaches considered inclined landing and none of the methods developed can be directly applied to this problem.

Inclined landing has been the topic of several works outside the deep learning control literature. A nonlinear hybrid controller was proposed in \cite{Dougherty2014LaserbasedGO}.
A trajectory-tracking controller first guides the quadrotor above the landing platform and then switches to an attitude-tracking controller to ensure that the attitude of the quadrotor adjusts to the slope of the landing platform upon touchdown. This is an ad hoc local strategy, incapable of generating optimal landing trajectories from arbitrary initial conditions. Besides, no real-time control experiments have been reported in this paper. The method proposed in \cite{inclined_landing_bebop_2015} features a nonlinear MPC to land a quadrotor on a moving inclined surface. Real-time experimental results were reported, showing a successful landing. The limitations of MPC are its computational complexity and the difficulty of parameter tuning, especially of the prediction horizon, which needs to be long for some of the landing trajectories, making the method unsuitable for embedded implementation on the quadrotor. The approach developed in \cite{Perching2016} relies on splitting up the problem in the generation of dynamically feasible trajectories and their subsequent trajectory tracking. Perching on slopes of up to 90 degrees has been demonstrated in lab experiments. To keep the problem tractable, the authors break the desired trajectory down into segments with a maximum duration of one second. The overall approach is more complex than the nonlinear feedback policy approach pursued in this paper.

\section{Simulation Model}\label{model}
The dynamic model of the Crazyflie 2.1 Nano-UAV is formed by the equations of motion (EOM). We divide them into the Newton-Euler equations, which govern the axial accelerations,  and an approximation of the body attitude control loops. The command input vector $u$ to the Crazyflie's onboard controller is defined as
\begin{equation}
\label{eq:InputvectorCrazyflie}
\begin{aligned}
u = \begin{bmatrix}
\Theta_{c} &
\phi_{c} & \theta_{c} & \dot{\psi}_{c}
\end{bmatrix}^{T}.
  \end{aligned}
\end{equation}
Here, $\Theta_{c}$ is the commanded pulse-width modulation (PWM) signal representing the total thrust, $\phi_{c}$ and $\theta_{c}$ are the commanded roll and pitch angles, respectively, and $\dot{\psi}_{c}$ is the commanded yaw rate \cite{crazyflie_ROS}. These inputs are bounded by
\begin{equation}
    \label{eq:Inputvectorbounds}
\begin{aligned}
u_{\min} &= \begin{bmatrix}
10000&
-30^{\circ} & -30^{\circ} & -200^{\circ}/s
\end{bmatrix}^{T},\\
u_{\max} &= \begin{bmatrix}
60000&
30^{\circ} & 30^{\circ} & 200^{\circ}/s
\end{bmatrix}^{T}.
  \end{aligned}
\end{equation}

\subsection{Newton-Euler Equations}

\begin{figure}[htbp]
    \centerline{\includegraphics[width=0.7\linewidth]{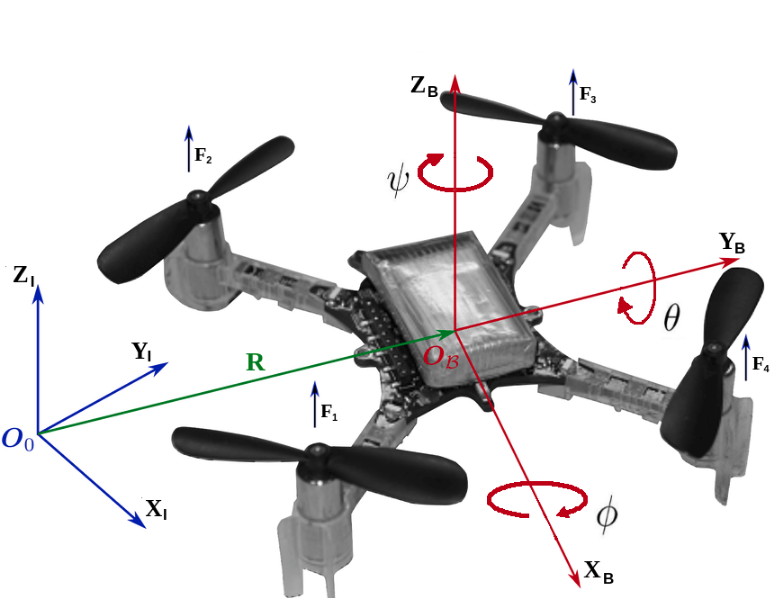}}
    \caption{The quadrotor coordinate system used throughout this paper. Subscripts B and I represent the body and inertial frame, respectively, and $F_{i}$ is the thrust due to rotor $i$. Adopted from \cite{crazyflieimage}.}
    \label{fig:axis}
\end{figure}

The quadrotor is modeled as a rigid body, with the axial accelerations in the inertial frame $\begin{bmatrix}x & y & z
\end{bmatrix}^{T}$:
\begin{equation}\label{eq:xyznewtoneuler}
  \begin{bmatrix}
    m\ddot x \\ m\ddot y \\ m\ddot z
  \end{bmatrix} =
  \textbf{R}
  \left(
  \begin{bmatrix}
    0\\
    0 \\
    F_{t}
  \end{bmatrix} +
    F_{a}
  \right)
  +
  \begin{bmatrix}
    0\\
    0 \\
    -mg
  \end{bmatrix}
\end{equation}
with $m$ the quadrotor's mass, \textbf{R} the rotation matrix from the body frame to the inertial frame, $F_{t}$ the total thrust force and $F_{a}$ the drag force. The rotation matrix corresponding to the coordinate frame representation in \figref{fig:axis} is
\begin{equation}
     \textbf{R} =
    \begin{bmatrix}
    c\psi c\theta - s\phi s\psi s\theta & -c\phi s\psi & c\psi s\theta + c\theta s\phi s\psi\\
    c\theta s\psi + c\psi s\phi s\theta & c\phi c\theta & s\psi s\theta + c\psi c\theta s\phi \\
    -c\phi s\theta & s\phi & c\phi c\theta
  \end{bmatrix}
  \end{equation}
where $c$ is a cosine, $s$ is a sine, and $\phi$, $\theta$ and $\psi$ are the roll, pitch and yaw angles, respectively. 

The relation between the commanded PWM\ $\Theta_{c}$ in \eqref{eq:InputvectorCrazyflie} and $F_{t}$ in \eqref{eq:xyznewtoneuler} is modelled by using the discrete-time transfer function found in \cite{Crazyflieident} for an individual motor $i$:
\begin{equation}
    \label{eq:thrustconversion500Hz}
\begin{aligned}
 \frac{F_{i}(z)}{\Theta_{c,i}(z)} = \frac{7.2345374\cdot 10^{-8}}{1-0.9695404z^{-1}} \qquad (500Hz).
  \end{aligned}
\end{equation}
Since the Crazyflie's onboard controller only takes a single PWM signal for all four motors, we assume that $F_{t} \approx 4F_{i}$ with $\Theta_{c} \approx \Theta_{c,i}$. Multiplying \eqref{eq:thrustconversion500Hz} by four and converting it to continuous-time gives
\begin{equation}\label{eq:thrustconversiontotal}
\begin{aligned}
  \begin{bmatrix}
    \dot{\Omega} \\
    F_{t}
  \end{bmatrix}& =
  \begin{bmatrix}
    -15.467\\
    1.425\cdot 10^{-4}  \end{bmatrix}\Omega +
  \begin{bmatrix}
    1\\
    2.894\cdot 10^{-7}
  \end{bmatrix}\Theta_{c}
  \end{aligned}
\end{equation}
where $\Omega$ is an unmeasured state used for simulation purposes only. The drag force $F_{a}$ in \eqref{eq:xyznewtoneuler} is expressed as \cite{Mueller2015}
\begin{equation}
    \label{eq:2DragForce}
\begin{aligned}
 F_{a} &= \boldsymbol{K}_{a} {\omega}_{\Sigma}v
  \end{aligned}
\end{equation}
with $\omega_{\Sigma}$ the sum of the rotor velocities, $v$ the body-frame velocity vector and $\boldsymbol{K}_{a}$ a diagonal matrix of drag constants estimated in \cite{Crazyflieident}. Because $\omega_{\Sigma}$ is not known during simulation, we approximate it from $F_{t}$ with additional conversion formula's given in \cite{Crazyflieident}.

\subsection{Body Attitude Control Loops}

The body attitude rates are modelled by equations that approximate the dynamics of the attitude control loops \cite{Greyboxalonsomora}:
\begin{equation}
    \label{eq:greyboxalonsomora}
\begin{aligned}
         \dot \phi &  = \frac{1}{\tau_{\phi}}(k_\phi\phi_c - \phi), \\
        \dot \theta & = \frac{1}{\tau_{\theta}}(k_\theta\theta_c - \theta), \\
        \dot \psi &  = \dot{\psi}_{c}.
\end{aligned}
\end{equation}
Here $\tau_{\phi}$, $\tau_{\theta}$ and $k_{\phi}$, $k_{\theta}$ are the time and gain constants for roll and pitch, respectively. The yaw rate is assumed to instantaneously track the desired yaw rate, which is a reasonable assumption since the yaw has no effect on the quadrotor's position \cite{Greyboxalonsomora}.
Because the closed-loop dynamics are unknown, the  parameters $k_{\theta}$, $k_{\phi}$, $\tau_{\theta}$ and $\tau_{\phi}$ need to be identified. Given the quadrotor's symmetry,  $k_{\theta}$ and $k_{\phi}$ as well as $\tau_{\theta}$ and $\tau_{\phi}$ are assumed equal.
These parameters are estimated by fitting the data gathered by a motion-capture system to the equations \eqref{eq:greyboxalonsomora}. We conducted 20 experiments using square and sine waves ranging from zero to thirty degrees, which gave an average fit of $85.3 \%$ using Matlab's \verb|nlgreyest| function, with the resulting parameters $k_{\phi}=k_{\theta} = 1.1094$ and $\tau_{\phi} = \tau_{\theta} = 0.1838$\,s.

To simulate the quadrotor, the model equations \eqref{eq:xyznewtoneuler} and \eqref{eq:greyboxalonsomora} are integrated by using the fourth-order Runge Kutta (RK4) method. The step size is fixed and equal to the sampling period $T_s = 0.02$\,s.

\section{Training Deep Reinforcement Learning Policies}\label{method}

To train the inclined landing, the quadrotor model of \secref{model} is used as a simulation environment for model-free DRL. Additionally, to navigate the quadrotor, we train set-point tracking in the same fashion. Both policy networks map quadrotor states to the desired control input in \eqref{eq:InputvectorCrazyflie}, which makes them directly applicable to the real Crazyflie. 

\subsection{Preliminaries}

The learning controller (agent) interacts with the model (environment) through trials. The environment's state space is denoted by $\mathcal{S}$ and a specific value of the state at time step $k$ by $s_{k}$. The agent applies an action $a_{k} \in \mathcal{A}$ and subsequently receives a reward $r_{k} \in \mathbb{R}$ , after which it observes the next state $s_{k+1}$. The action $a_{k}$ is chosen by following a stochastic policy $\pi_{k}(a|s)$ or a deterministic policy $\mu_{k}(s)$. This policy can be optimized in many different ways. Most techniques maximize the discounted return $\eta(\pi_{\phi}) = \EX _{\tau} [\sum_{t=0}^{T} \gamma^{t}r(s_{k},a_{k})]$, with $\tau$ a trajectory following the policy $\pi_{\phi}$ and $\gamma$ the discount factor.

\subsection{Set-Point Tracking}
We first train a three-dimensional set-point tracking policy network to empirically check the simulation-to-reality performance of the DRL algorithms and to fly to a desired starting position for inclined landing. For this task, the states and actions are defined as follows:
\begin{equation}
    \label{eq:3dvectorsim}
\begin{aligned}
s_{3d} &= \begin{bmatrix}
x & y & z & v_{x} & v_{y} & v_{z} &\phi & \theta
\end{bmatrix}^{T},                                                                \\
a_{3d} &= \begin{bmatrix}
\Theta_{c} & \phi_{c} & \theta_{c}
\end{bmatrix}^{T}.
  \end{aligned}
\end{equation}
Here, $v$ is the velocity in the inertial frame. Note that the yaw angle is kept constant at zero degrees and can thus be omitted throughout all our experiments. For set-point tracking, we use the following reward function:
\begin{equation} \label{eq:euclideanreward}
  r_{k} = -e_{p} -0.2e_{v} -0.1e_{\phi,\theta} -0.1\frac{a_{\phi,\theta}^{2}}{\max(e_{p}, 0.001)}
\end{equation}
where $e_{p}$, $e_{v}$ and $e_{\phi,\theta}$ are Euclidean distance errors of the position, velocity, and orientation, respectively, with respect to the goal state. The term $a_{\phi,\theta}^{2}$ is the sum of the squared roll and pitch actions (normalized between $0$ and $1$). It is scaled by the reciprocal of $e_{p}$ to minimize oscillations near the goal position.

The policy network is a fully connected neural network with two hidden layers, with 64 neurons each, and the tanh activation function everywhere except for the output layer which has a linear activation function. The final output is subsequently clipped between $-1$ and $1$. We use the PPO algorithm \cite{PPOSchulmanWDRK17} to train the set-point tracking network. Other state-of-the-art DRL algorithms like twin delayed deep deterministic policy gradient (TD3) \cite{TD3} and soft actor critic (SAC) \cite{SAC} converged successfully as well, but PPO was superior in terms of training time and final policy performance.

\subsection{Inclined Landing}

To keep the state dimensions small and the DRL problem tractable, the inclined landing is trained in the $xz$-plane, representing the three-dimensional model restricted to a two-dimensional plane. The states and actions used are :
\begin{equation}
    \label{eq:2dvectorsim}
\begin{aligned}
s_{2d} &= \begin{bmatrix}
x & z & v_{x} & v_{z} & \theta
\end{bmatrix}^{T},                                                                \\
a_{2d} &= \begin{bmatrix}
\Theta_{c} & \theta_{c}
\end{bmatrix}^{T}.
  \end{aligned}
\end{equation}
For the sake of brevity, in the sequel, we refer to $s_{2d}$ and $a_{2d}$ by $s$ and $a$, respectively.
Because the quadrotor is under-actuated, an initial swinging motion away from the landing location is required for some initial conditions. This characteristic is incompatible with the bias of a Euclidean distance based reward like the one in \eqref{eq:euclideanreward} generates. The reward function used for the inclined landing is therefore a sparse reward defined as follows:
\begin{equation} \label{eq:2dsparsereward}
  r_{k} =
  \begin{cases}
     0& \text{if} \quad s_{k} \in S_{g}\\
    -\beta & \text{if} \quad s_{k} \in S_{o} \\
    -2& \text{if} \quad s_{k} \in S_{b} \\
    -1 & \text{otherwise.}
  \end{cases}
\end{equation}
%
Here $S_{g} = \{s \;\vert\; |s_i - s_{g,i}| < \delta_{g,i}, \, \forall i \}$ is the set of goal states, defined as a hyperbox around the landing attitude. The goal threshold vector $\delta_{g}$ defines the desired landing tolerance around the goal state $s_{g,i}$ and is set by the user.
The landing platform itself is an obstacle associated with a set of obstacle states $S_{o}$ and a penalty $\beta$, and $S_{b}$ represents the set of states close to the state space boundaries.

\begin{figure}[htbp]
    \centerline{\includegraphics[width=.9\linewidth]{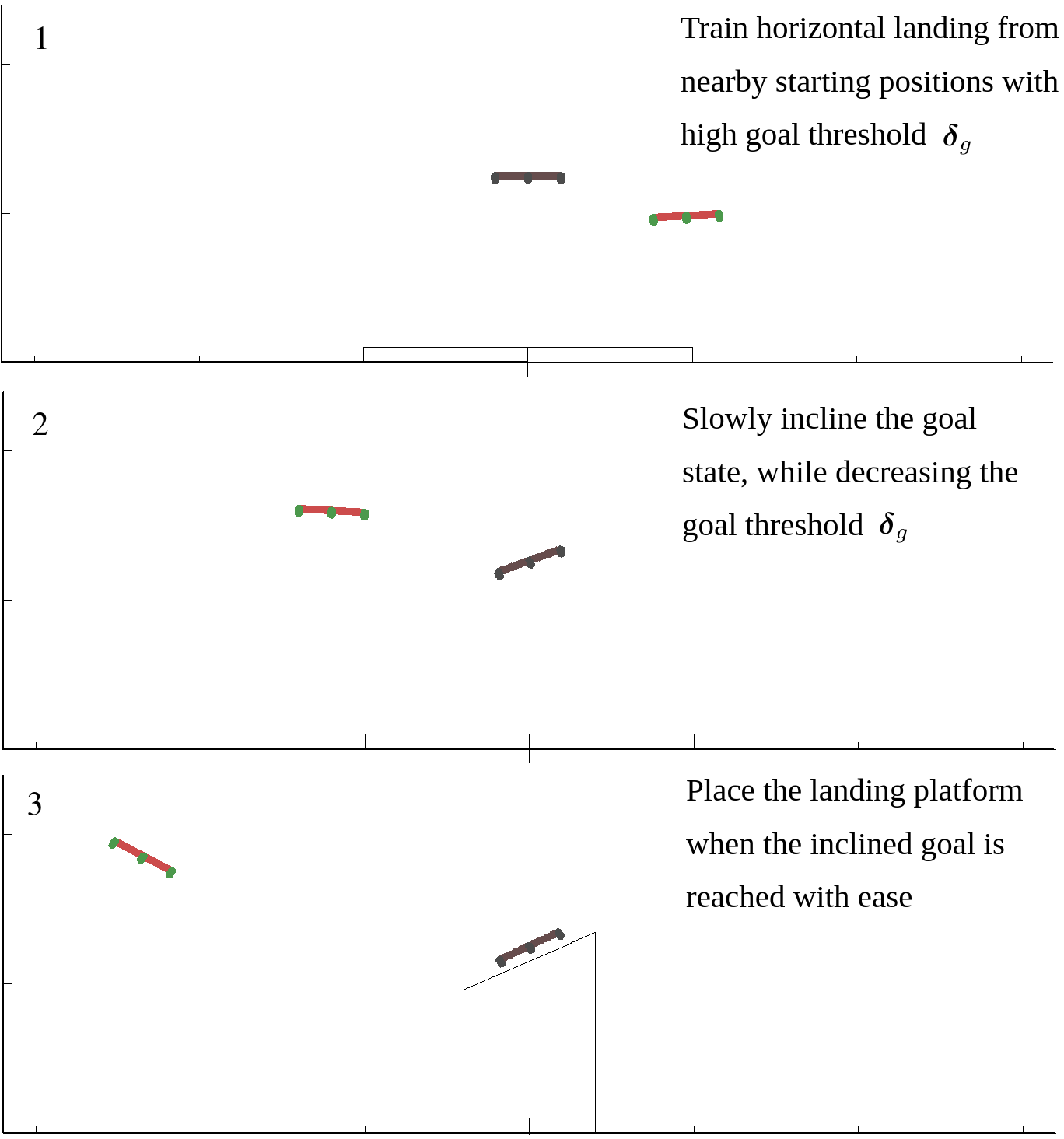}}
    \caption{The progress of the training curriculum for inclined landing. The red quadrotor represents the agent and the black quadrotor represents the goal state.}
    \label{fig:progressions}
\end{figure}

The use of a sparse reward requires extensive exploration to receive a non-negative reward and often leads to prolonged or unsuccessful training. We introduce the following curriculum learning \cite{curriculum} procedure to speed up the training and achieve convergence:
\begin{itemize}
\item The training starts without a landing platform and with a horizontal goal state. Only once the quadrotor reliably reaches the horizontal goal, we begin slightly tilting the goal position after each episode. Finally, the landing platform is introduced into the environment, see \figref{fig:progressions}.
\item We initialize simulations near the goal state and with each episode expand the set of initial positions. This eliminates the need for exploration by letting the non-negative rewards propagate throughout the value network at the beginning of training.
\item We start with a large goal hyperbox $S_{g}$ and as the training progresses, the hyperbox is gradually reduced to its desired size.

\end{itemize}
This learning curriculum requires an on-policy learning algorithm, such as PPO. Off-policy replay buffers would inevitably contain samples representing goals that are no longer relevant. In our experience, off-policy algorithms TD3 and SAC cannot keep up with the curriculum. The policy network architecture is similar to the one used for set-point tracking. The input and output layers for inclined landing are the state and action vectors in \eqref{eq:2dvectorsim}.

\subsection{Simulation to Reality Transfer}

\begin{figure}[htbp]
    \includegraphics[width=1\linewidth]{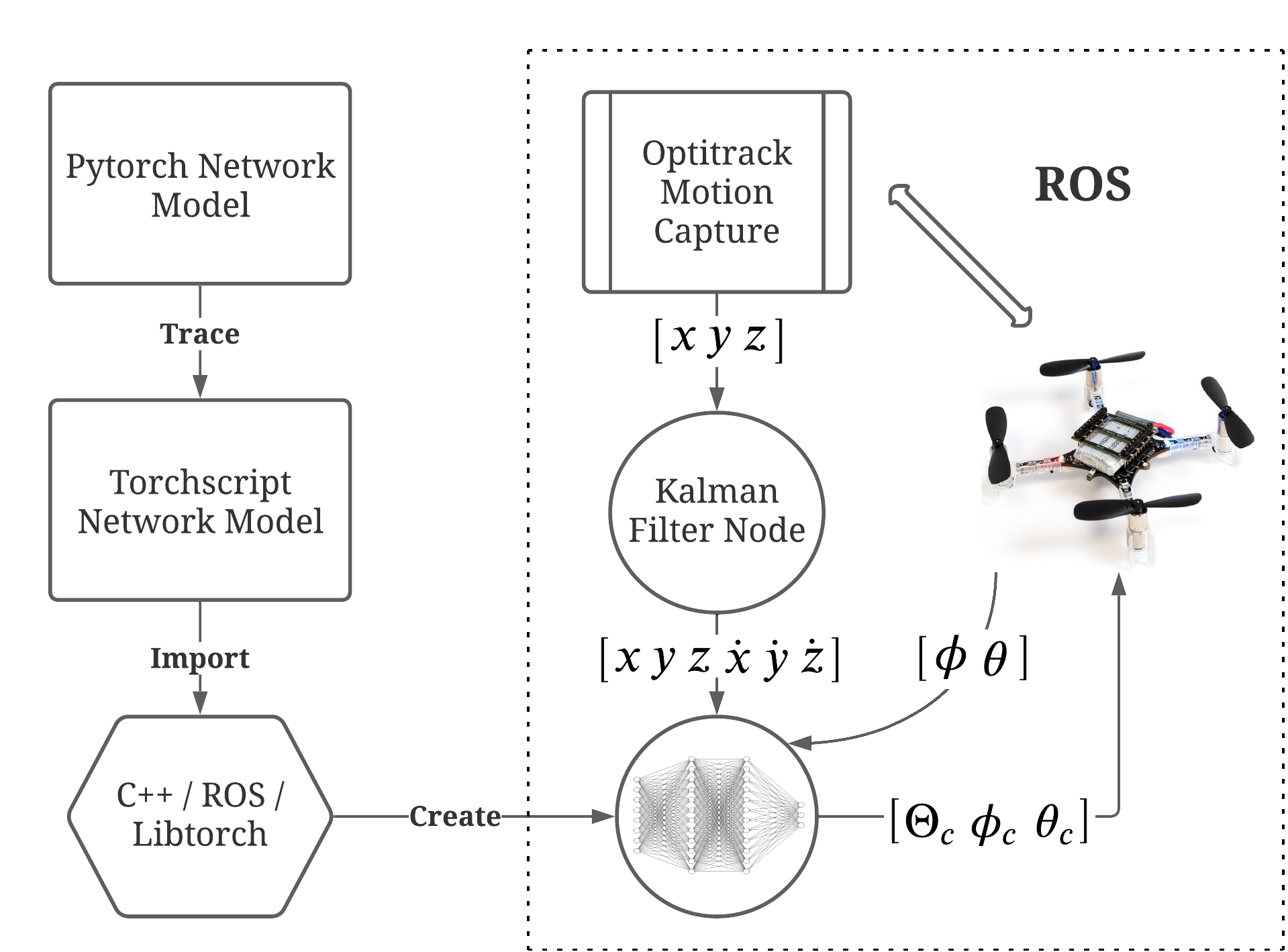}
    \caption{Schematic overview of the Pytorch model deployment.}
\label{fig:S2R}
\end{figure}

To deploy the policies on the Crazyflie 2.1 Nano-UAV, the trained Pytorch network is converted to work within a Robot Operating System (ROS) node, which maps the state to the control input vector in \eqref{eq:InputvectorCrazyflie}. The positions and velocities come from a Kalman filter node, which appends the coordinates from the Optitrack motion capture system with the estimated inertial frame velocities. The quadrotor's orientation is taken from the Crazyflie's default onboard estimator. An overview of this process is given in \figref{fig:S2R}.

\section{Experiments}\label{experiments}

All simulations, DRL training, and lab experiments are done on an HP Zbook Studio G4 laptop, with the default Nvidia Quadro M1200 GPU and an Intel Core i7-7700HQ CPU. The additional hardware used is the Optitrack motion-capture system, the standard Crazyflie 2.1 with a small marker holder, and a Crazyradio PA for communication with the Crazyflie.

\subsection{Experimental Setup}

The simulations and DRL\ training are implemented in Python, using the Pytorch package. The quadrotor dynamics are simulated using the EOM of \secref{model}, integrated by the RK4 method. To increase the integration speed, we compile the EOM functions using Numba's Just-in-time (jit) package. By keeping the simulations and rendering in our own Gym-architecture \cite{Gym} environment, we can use well-tuned implementations of existing model-free algorithms by \cite{stable-baselines3}. Our compact simulator allows for computationally cheap rendering, easy curriculum adjustments and fast simulation. The simulation bounds are equal to the dimensions of the actual flying arena with
$
\begin{bmatrix}
  x_{min}&x_{max}&y_{min}&y_{max}&z_{min}&z_{max}\end{bmatrix} = \begin{bmatrix}
  -3.4&3.4&-1.4&1.4&0&2.4\end{bmatrix}$\,m.
The actions represent the multiplication of the clipped policy network outputs $(a \in [-1,1])$ by the control bounds in \eqref{eq:Inputvectorbounds}, with an exception for the PWM network output which is multiplied by $16500$ and added to the estimated hover PWM of $42000$.
 All simulations use the control sampling frequency of $50$\,Hz, with episode length of 300 time steps, i.e., 6 seconds.

\begin{figure*}
    \centering
    \includegraphics[width=1\textwidth]{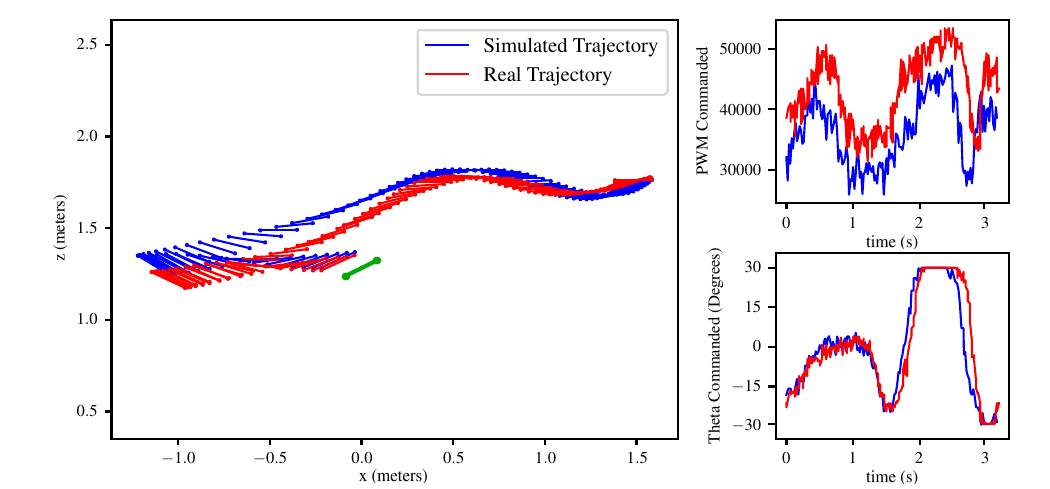}
\end{figure*}

\begin{figure*}
    \centering
    \includegraphics[width=1\textwidth]{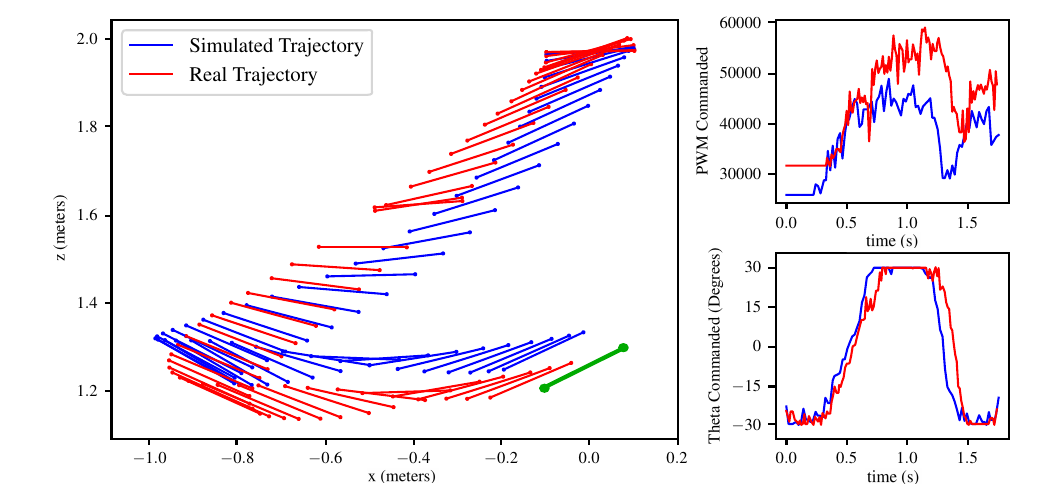}
    \caption{Landing trajectories starting from above right (top) and from directly above (bottom) of the inclined goal position (green) at (0, 1.25). The trajectory ends when the quadrotor is within the set of goal states $S_{g}$.}
    \label{trajectories}
\end{figure*}

\subsubsection{Set-Point Simulations}

The goal state is taken as $s_{g}= \begin{bmatrix}0 & 0 & 1.2 & 0 & 0 & 0 & 0 & 0
\end{bmatrix}^{T}$, which represents the center of our physical flying arena. After each episode, the initial position is set randomly anywhere within the simulation bounds, with a small margin around the edges. The PPO implementation of \cite{stable-baselines3} is used, with the discount factor $\gamma$ reduced to $0.97$ to account for more short-term control behaviour.  We train for a total of $10^{6}$ time steps representing 27 minutes of training time. For navigation with the resulting policy, a coordinate change suffices to fly the quadrotor to an arbitrary position.

\subsubsection{Inclined Landing Simulations}

For inclined landing, we operate the quadrotor in the $xz$-plane.
The DRL algorithm used is the PPO implementation of \cite{stable-baselines3}, where minor changes have been made to activate the rendering every 50 training iterations and to gradually start increasing $\gamma$ from $0.97$ to $0.99$ after 300 training iterations. An episode ends at 300 time steps or when the state $x$ is within the goal hyperbox $S_{g}$. The landing platform is modeled as a polygon and appears after $8\cdot 10^{5}$ time steps. The goal threshold vector is defined as $\delta_{g} = \begin{bmatrix}
 \delta_{x}  & \delta_{z} & \delta_{v_{x}} & \delta_{v_{z}} & \delta_{\theta}
\end{bmatrix} = \begin{bmatrix}
d  & d & \min(10d,1.5) & \min(10d,1.5) & 0.25d
\end{bmatrix}$, where $d$ starts at $0.25$\, in the beginning of training and gradually decreases to $0.10$ after $2500$ episodes with $0.15/5000$ per episode. The box of possible starting positions around the goal state expands with every episode by $1/6000$\,m in the $x$-direction and by $1/8000$\,m in the $y$-direction. Additionally, the goal state stays horizontal for the first $4\cdot 10^{5}$ time steps and then gradually tilts towards its final inclination of $-\pi /7$ at the rate of $(-\pi /7)/6000$ radians per episode. The final goal state is set at $s_{g}= \begin{bmatrix}0 & 1.25 & 0 & 0 & -\pi /7
\end{bmatrix}^{T}$. The obstacle reward constant $\beta$ is taken as $-7$, which was found empirically to be the right trade-off between the goal of landing on top of the platform and the necessity to avoid the landing platform base. Training is stopped anywhere between $1.2\cdot 10^{6}$ and $3\cdot 10^{6}$ time steps (30 to 80 minutes), when the rendering shows that the policy executes the inclined landing reliably. Note that rather than monitoring the loss function, the aforementioned parameters and curricula have been empirically tuned by frequently rendering. The trend in the loss function value is quite meaningless, given the curricula and the discount factor adaptation.

\subsubsection{Validation on a real quadrotor} The quadrotor used is the Crazyflie 2.1 Nano-UAV with its original firmware. A Crazyflie-specific package \cite{crazyflie_ROS} allows us to publish the control values in \eqref{eq:InputvectorCrazyflie} directly to the ROS server, which are then transmitted with low latency to the Crazyflie over the Crazyradio PA. The trained policy network is evaluated at $80$\,Hz, even though it was trained at $50$\,Hz. This is possible, as the policy is a function of the physical state only, and can therefore be evaluated at any  frequency. A single policy evaluation takes approximately $2.5$\,ms. The coordinates from the Optitrack motion-capture system come in at $120$\,Hz and are combined with the inertial frame velocities by the Kalman filter node \cite{Zhu2019RAL}. The orientation is received from the Crazyflie's onboard estimator through the Crazyradio at around $80$\,Hz. The position, velocity, and orientation estimates form the policy input. The experiment itself starts with the drone flying to a position of choice, using the three-dimensional set-point tracking network. Once it is positioned, the networks are switched and the drone commences the inclined landing. The initial state can be at an arbitrary location in the top half of the flying arena. The landing trial ends when the quadrotor's state $s_{k}$ reaches the set of goal states $S_{g}$.

\subsection{Experiment Results}

Early experiments showed a vertical offset between the simulated and real trajectories, caused by a slight inaccuracy of the motor thrust equation \eqref{eq:thrustconversiontotal}. We compensated for this offset by increasing the hover PWM to 48000 during the flying arena testing. The resulting behaviour was very similar to the simulations, as can be seen in \figref{trajectories}.\footnote[5]{Because of a faulty Optitrack measurement, the bottom figure shows a single misplaced red quad with a corresponding short drop in the PWM output.} These trajectories originate from the same policy network, starting from arbitrary initial positions and ending when $s_{k} \in S_{g}$ with $\delta_{g} = \begin{bmatrix}
  0.10 & 0.10 & 1.5 & 1.5 & 0.025
\end{bmatrix}$.

To further evaluate the performance, we measured the landing success rate when starting the flight from three different initial positions. Each position was evaluated $10$ times, resulting in the success percentages reported in Table \ref{table_success}. Even if an experiment did not succeed, the agent would not crash but autonomously fly back and forth, sometimes succeeding in its second or third attempt. However, these further attempts were not counted as a successful landing.

\begin{table}[h]
\caption{Comparison of Real-World and Simulation Experiments}
\label{table_success}
\begin{center}
\setlength\tabcolsep{4.1pt}
\begin{tabular}[t]{l| c c c | c c} 
\hline setting &  & success from initial $(x, z)$ & &\textbf{total} \\
\hline \hline 
 &  $(0, 2)$ & $(-1.5, 1.6)$ & $(1.5, 1.8)$ & \\
\hline
Real-World & 90\%  & 70\% & 100\% & \textbf{86.7}\textbf{\%}  \\
Simulation & 90\% & 90\%  & 100\%  & \textbf{93.3}\textbf{\%} \\
\hline
\end{tabular}
\end{center}
\end{table}
No simulation-to-reality transfer techniques, such as domain randomization, have been employed, as they did not seem to improve the final performance, while they did complicate the agent's training. Additionally, we found that using the Crazyflie's onboard orientation estimates rather than the Optitrack orientation estimates resulted in a substantial improvement in the consistency of performance.

The results show that inclined landing controllers can be designed by means of DRL. These controllers can transfer adequately to reality without the need for dynamics randomization, or sensor noise. Furthermore, the same policy can be executed from a wide variety of initial states. The performance could further improve by additional system identification of the total motor thrust. The slight mismatch observed might have been caused by making too strong assumptions about the scaling of the single motor model in \eqref{eq:thrustconversion500Hz} to the full motor model in \eqref{eq:thrustconversiontotal}. However, a perfect thrust model will never exist, due to the motor wear and tear and the strong relation of the drone's battery level to its thrust output, which was also reported in \cite{QuadRL2017Hwangbo}.

\section{DISCUSSION}
\label{discussion}

\subsection{Onboard Controller Dynamics}

Using a model that incorporates the inner-loop system dynamics of the Crazyflie's onboard controllers allowed us to fly the quadrotor by using intermediate-level control commands. Although this requires a quadrotor-specific parameter identification, the procedure was straightforward (as also reported in \cite{Greyboxalonsomora}), took a negligible amount of time, and can be applied to any onboard-controlled quadrotor.

Furthermore, quadrotor end-users usually prefer to keep the onboard controllers in place, as they provide basic functionality and safety features. This has therefore been the main reason we conducted the research in this setting. A drawback of this approach compared to training DRL on individual motor thrusts \cite{QuadRL2017Hwangbo,Honigdeeprlcrazyflie} has been the  limit of $30^{\circ}$ of the attitude controller, restricting us to maximum landing angles of around $25.7^{\circ}$. However, the benefit of using the closed-loop model is that the policy does not need to stabilize the quadrotor in the first place, and can focus on using the quadrotor's stable dynamics to learn the behaviour needed for the inclined landing.

\subsection{DRL Algorithms and Curricula}
For set-point tracking, PPO was far superior to TD3 or SAC in terms of performance and training time. For inclined landing, we have tested these off-policy algorithms with reduced replay buffers to cope with the changing goal state (between $1\cdot 10^{4}$ and $3\cdot 10^{4}$ samples). The resulting policies were very poor, shaky and not capable of yielding a smooth landing behaviour in the last curriculum phase.

In the beginning of training, having a larger goal threshold $\delta_{g}$ was important for convergence, even when starting at and around the goal state. Having a too large goal velocity threshold would however cause strong oscillations during training, hence the minimum term in the training value for $\delta_{v_{x}}$ and $\delta_{v_{z}}$. The convergence during the subsequent inclined landing phase was mostly dependent on the performance during the horizontal phase. As long as the quadrotor could adequately reach the horizontal hover position, it would be able to transition into larger angles by means of small increments. Interestingly, this is where the quadrotor learns the swinging behaviour on its own, making use of its inherent dynamics. The success of the final phase depends on the obstacle penalty $\beta$, where a small value will make the agent exploit the platform for braking, and a large value will diminish the incentive to reach the goal state. The final curriculum works for landing angles of $25.7^{\circ}$ with an attitude controller limitation of $30^{\circ}$, but simulations using a hypothetical attitude controller limitation of $55^{\circ}$ have showed that angles of $50^{\circ}$ can also be reached within the same time-span and using the same training procedure.

\subsection{Landing in the vertical xz-plane}

Because the landing controller is planar, it  will not compensate drift in the y-axis. Since our experiments were conducted indoors, this problem was negligible. However, when conducting experiments under disturbances or with poorly calibrated quadrotors, one could add a basic PID controller regulating the alignment with the y-axis.

We believe that the planar controller is still very applicable in real settings. For inclined landings, as we envisage them, the landing platform will almost always be approachable in a plane (exceptions would be strong disturbances like side wind or complicated obstacles in the approach trajectory).  This means that a quadrotor can fly toward the platform, change its yaw axis accordingly, and perform the inclined landing using the planar controller.

\section{CONCLUSION AND FUTURE WORK}\label{reality}

We have presented a model-free DRL technique to facilitate autonomous quadrotor landing on an inclined surface. We trained a control agent with PPO, using sparse rewards and a learning curriculum. This allows the agent to gradually progress toward a more difficult task, i.e., larger inclination angles of the landing platform. Moreover, we have shown that the trained policies transfer well to reality, without employing any simulation-to-reality transfer techniques.

A limitation of this work is the fact that we restricted the landing trajectory to the xz-plane, which may cause some drift in the y-direction. A three-dimensional landing policy could increase precision, albeit at the cost of longer and more complex training. 

An extension to such a setting is the topic of our future work.
Preliminary simulation results show that training a three-dimensional policy is feasible and converges within a similar time span as the two-dimensional policy.
Additional future work could focus on larger inclination angles of the landing platform, as long as the onboard attitude controller would allow them. In this way, one could try to extend the findings in this paper to a perching behavior similar to \cite{Perching2016}. The goal inclination can also be added to the quadrotor's state, which would enable landing on unknown platforms, using, for instance, a laser measurement system \cite{Dougherty2014LaserbasedGO} or an onboard camera system \cite{inclined_landing_bebop_2015}. Another interesting path for future work is to train a single inclined landing policy that is applicable on multiple quadrotors with comparable onboard control architectures, similar to the work in \cite{Honigdeeprlcrazyflie}. Finally, a form of the platform contact dynamics in \cite{landing_multirotor} could be implemented in our simulator for a more robust landing and to aid the design of an end-to-end controller, which would eliminate the need for an external stopping signal.




\section*{ACKNOWLEDGMENT}
We thank our colleague Hai Zhu for his guidance and advice in the initial stage of this project.
Robert Babu\v{s}ka  was supported by the European Union's H2020 project Open Deep Learning Toolkit for Robotics (OpenDR) under grant agreement No. 871449.

\bibliographystyle{IEEEtran}
\balance
\bibliography{root}

\end{document}